# Determining Determiner Sequencing: A Syntactic Analysis for English

B.A. Hockey
Dept. of Linguistics
University of Pennsylvania
Philadelphia PA 19104-6305
email: beth@linc.cis.upenn.edu

D. Egedi
Institute for Research in Cognitive Science
University of Pennsylvania
Philadelphia PA 19104-6228
email: egedi@linc.cis.upenn.edu

cmp-lg/9411004  3 Nov 94

## Abstract

Previous work on English determiners has primarily concentrated on their semantics or scoping properties rather than their complex ordering behavior. The little work that has been done on determiner ordering generally splits determiners into three subcategories. However, this small number of categories does not capture the finer distinctions necessary to correctly order determiners. This paper presents a syntactic account of determiner sequencing based on eight independently identified semantic features. Complex determiners, such as genitives, partitives, and determiner modifying adverbials, are also presented.

## 1  Introduction

Most work on determiners has been concerned with either purely semantic properties, the occurrence of particular determiners in certain syntactic environments such as existential-*there* sentences, determiners as heads of phrases (the DP hypothesis), or quantifier scoping.

One question that has not been extensively discussed is how the class of determiners order with respect to each other. We have identified a set of determiner features motivated solely by semantic properties [4, 6] that when used together have the syntactic consequence of accounting for the complex patterns of determiner sequencing. Although we do not claim to have exhaustively covered the rich determiner system of English, we do cover a large subset, both in terms of the phenomena handled and in terms of corpus coverage. We were able to find 103 lexical items marked as determiners in the Collins English Dictionary. This analysis accounts for 82 of these. Of the 21 remaining, we believe that at least 8 should be marked as adjectives. A discussion of the remaining determiners occurs in the Future Work section of this paper. Most of the analysis presented in this paper is implemented as part of a Feature-Based, Lexicalized Tree Adjoining Grammar (FB-LTAG) for English [3, 5, 8].

Previous approaches to syntactic determiner ordering (e.g. [7]) have simply divided determiners into subcategories (e.g. predet, det, postdet). This type of approach is inadequate because it allows ungrammatical sequences like *all what no*, and misses the finer distinctions among particular determiners. These finer distinctions are modeled very naturally in a lexicalized grammar formalism such as FB-LTAG in which pieces of syntactic structure and features representing linguistic properties are associated with individual lexical items.

In our account of determiner sequencing, there is a set of eight features that form the core of the system. Each determiner carries with it a set of values for these features that represents its properties, and a set of values for the properties of any determiners it may modify. Various constructions, such as the number system, genitives, and partitives, as well as the adverbs that interact with the determiner system, use these features to indicate which determiners can participate in the constructions, and which determiners can be modified.

## 2  Determiner Ordering Using Features

In our English FB-LTAG grammar, there are two kinds of basic noun phrases (NP), those that take determiners and those that do not. Nouns that take (or require) determiners[1] have a DetP substitution site. Complex DetP's are formed by having determiners adjoin onto each other. There are two basic determiner trees: an initial tree and an auxiliary tree. Figure 1 shows the initial and auxiliary trees as anchored by the determiner *these*. Since any single determiner can function as a complete DetP[2], every determiner selects the initial tree. Determiners that can modify other determiners also select the auxiliary tree.

The decision to include a DetP substitution node in the NP, as we have done, or to have the determiners adjoin on, as proposed in [1], is independent of this analysis of determiner ordering. We will not address the advantages and disadvantages of these two possi-bilities here. Suffice it to say that the general analysis will work with either approach, although of course certain trees would have to be changed. On a further note, we believe that this sequencing analysis is a general one, not specifically related to TAGs, and would work for any other feature-based formalism, assuming appropriate, formalism-specific modifications.

---

[1] Henceforth DetP's, not to be confused with 'DP's' as in the DP Hypothesis.

[2] By definition. Our main criteria in classifying something as a determiner was that it be able to stand alone with a noun to form an NP.

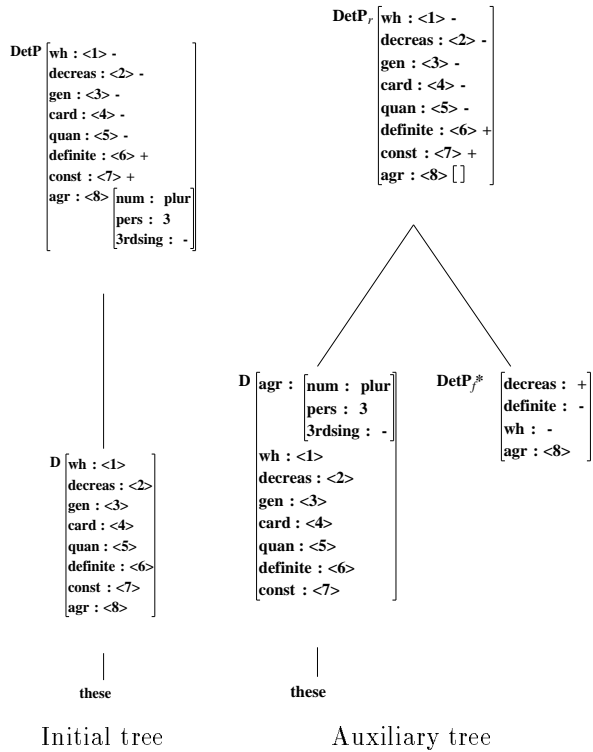

Figure 1: Determiner Trees with Features

| Det | defin | quan | card | gen | wh | decreas | const | agr |
|---|---|---|---|---|---|---|---|---|
| all | + | + | − | − | − | − | + | 3pl |
| this | + | − | − | − | − | − | + | 3sg |
| that | + | − | − | − | − | − | + | 3sg |
| what | + | − | − | − | + | − | + | 3sgpl |
| the | + | − | − | − | − | − | + | 3sgpl |
| every | + | + | − | − | − | − | + | 3sg |
| each | + | + | − | − | − | − | + | 3sg |
| any | − | + | − | − | − | − | + | 3sg |
| a | − | + | − | − | − | − | + | 3sg |
| no | + | + | − | − | − | − | + | 3sgpl |
| few | − | + | − | − | − | + | − | 3pl |
| many | − | + | − | − | − | − | − | 3pl |
| GEN | + | − | − | + | − | − | + | |
| CARD | + | + | + | − | − | − | + | 3pl[3] |
| PART | − | − | − | − | − | − | + | |

Table 1: Determiner Features

### 2.1 Identifying the features

In our analysis, features are crucial to ordering determiners correctly. We have identified eight features which are sufficient to order the determiners. These features are: **definiteness, quantity, cardinality, genitive, decreasing, constancy, wh**, and **agr**. These features have all been previously proposed as semantic properties of determiners. The semantic definitions underlying the features are given below.

**Definiteness:** Possible Values [+/−].
A function f is definite iff f is non-trivial and whenever $f(s) \neq \emptyset$ then it is always the intersection of one or more individuals. [4]

**Quantity:** Possible Values [+/−].
If A and B are sets denoting an NP and associated predicate, respectively; E is a domain in a model M, and F is a bijection from $M_1$ to $M_2$, then we say that a determiner satisfies the constraint of quantity if $Det_{E_1} AB \leftrightarrow Det_{E_2} F(A)F(B)$. [6]

**Cardinality:** Possible Values [+/−].
A determiner D is cardinal iff D ∈ cardinal numbers ≥ 1.

**Genitive:** Possible Values [+/−].
Possessive pronouns and the possessive morpheme (*'s*) are marked **gen+**; all other nouns are **gen−**.

**Decreasing:** Possible Values [+/−].
A set of Q properties is decreasing iff whenever $s \leq t$ and $t \in Q$ then $s \in Q$. A function f is decreasing iff for all properties f(s) is a decreasing set.

A non-trivial NP (one with a Det) is decreasing iff its denotation in any model is decreasing. [4]

**Constancy:** Possible Values [+/−].
If A and B are sets denoting an NP and associated predicate, respectively, and E is a domain, then we say that a determiner displays constancy if $(A \cup B) \subseteq E \subseteq E'$ then $Det_E AB \leftrightarrow Det_{E'} AB$. Modified from [6]

**Wh:** Possible Values [+/−].
Interrogative determiners are **wh+**; all other determiners are **wh−**.

**Agreement:** Possible Values [3sg, 3pl, 3sgpl].
Although English does not have the morphological marking of determiners for case, gender or number, we hold that most determiners in English are semantically marked for number.

The initial determiner tree in Figure 1 shows the appropriate feature values for the determiner *these*, while Table 1 shows the corresponding feature values of several other common determiners.

In addition to the features that represent their own properties, determiners that select the auxiliary tree have features to represent the selectional restrictions these determiners impose on the determiners they modify. The selectional restriction features of a determiner appear on the DetP footnode of the auxiliary tree that the determiner anchors. The $DetP_f$ node in the auxiliary tree in Figure 1 shows the selectional feature restriction imposed by *these*[4], while Table 2 shows the corresponding selectional feature restrictions of several other determiners.

### 2.2 Wh and Agr Features

A determiner with a **wh+** feature is always the leftmost determiner since no determiners can adjoin onto it. The presence of a wh+ determiner makes the entire

---
[3] except *one* which is 3sg
[4] In addition to this tree, *these* would also anchor another auxiliary tree that adjoins onto **card+** determiners.

| Det | defin | quan | card | gen | wh | decreas | const | agr |
|---|---|---|---|---|---|---|---|---|
| all | + | − | − | | − | | | |
| this | − | | + | | − | | + | |
| that | − | | + | | − | | + | |
| what | − | | + | | + | | | |
| the | − | | + | | − | | | − |
| every | − | | + | | − | | + | |
| each | − | | + | | − | | + | |
| any | − | | + | | − | | + | |
| a | − | | + | | − | | + | |
| many | | | | only nouns | | | | |
| no | | | | only nouns | | | | |
| GEN | | | | only nouns | | | | |
| CARD | | | | only nouns | | | | |
| PART | | | | | | | − | |

Table 2: Selectional Restrictions Imposed by Determiners

NP wh+, so this feature is always passed through to the NP node, unlike other features which are considered internal to the determiner system.

The **agr** feature functions differently from most of the features in the determiner sequencing system. Notice that in the auxiliary tree in Figure 1, the **agr** feature is the only feature not passed from the D node to the root DetP node, but is passed instead from the foot DetP to the root DetP. In the determiner system, the **agr** feature is generally propagated from the rightmost determiner (i.e. the one closest to the noun), although some adjoining determiners require that they also agree with that determiner. This distinction is captured in FB-LTAG by having each determiner specify in its lexical entry whether or not its agreement feature is passed to the root DetP (i.e. from the D node to the DetP$_r$ node).

## 3 Genitive Constructions

There are two kinds of genitive constructions: genitive pronouns, and genitive NP's (which have an explicit genitive marker, 's, associated with them). It is clear from examples such as *her dog was run over* vs *∗dog was run over* that genitive pronouns function as determiners and as such, they sequence with the rest of the determiners. The features for the genitives are the same as for other determiners. No **agr** is specified in Table 1, since the **number** and **person** of the genitive will depend on its particular form (i.e. *my* vs *their*). Genitives are not required to agree with either the determiners or the nouns that they modify.

Genitive NP's are particularly interesting because they are potentially recursive structures. Complex NP's can easily be embedded in a determiner phrase.

(1) [[[John]'s friend from high school]'s uncle]'s mother came to town.

There are two things to note in the above example. One is that in embedded NP's, the genitive morpheme comes at the end of the NP phrase, even if the head of the NP is at the beginning of the phrase. The other is that the determiner of an embedded NP can also be a genitive NP, hence the possibility of recursive structures.

In the FB-LTAG grammar, the genitive marker, 's, is separated from the lexical item that it is attached to and given its own category (G). In this way, we can allow the full complexity of NP's to come from the existing NP system, including any recursive structures. The two trees in Figure 2 demonstrate how easily the complexity of genitive NP's can be captured in FB-LTAG. As with the standard determiner trees, there are two trees - one for the determiner that stands alone and one for a determiner that adjoins onto another.

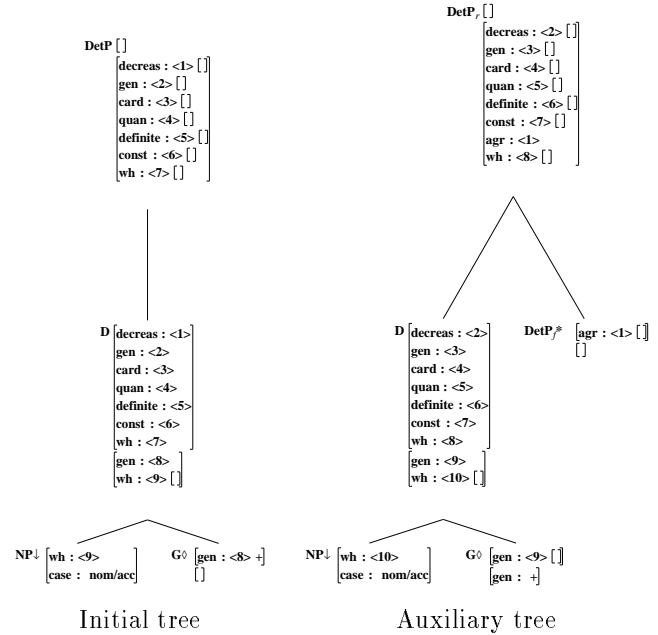

Figure 2: Genitive Determiner Trees

## 4 Partitive Constructions

The deciding factor for including partitive constructions (e.g. *some kind of, all of*) in the category of determiner constructions was the behavior of the agreement features. If partitive constructions are analyzed as an NP with an adjoined PP, then we would expect to get agreement with the head of the NP (as in Sentence (2)). If, on the other hand, we analyze them as a determiner construction, then we would expect to get agreement with the noun that the determiner phrase modifies (as we do in Sentence (3)).

(2) a *kind* [of these machines] *is* prone to failure.

(3) [a kind of] these *machines are* prone to failure.

Note that both the partitive and PP readings are possible for the same NP. The distinguishing characteristic is with which noun agreement occurs.

Partitive constructions, of course, interact with other determiners. Since they can modify the noun itself ('*[a certain kind of] machine*'), or modify other determiners ('*[some parts of] these machines*', the partitive construction has both an initial and auxiliary tree that are anchored by the preposition *of*.

## 5 Determiner Adverbs

There are some adverbs that interact with the NP and determiner system [7], although there is some debate in the literature as to whether these should be classified as determiners or adverbs. Sentences (4) - (6) contain examples of this phenomena.

(4) Hardly any attempt was made at restitution.

(5) Only Albert would say such a thing.

(6) Almost all the people had left by 5pm.

Adverbs that modify NP's or determiners have restrictions on what types of NP's or determiners they can modify. They divide into three classes based on the pattern of these restrictions. The adverbs *especially*, *even*, *just*, and *only* form a class that can modify any NP that is **wh–**, including Proper Nouns. A second class, consisting of adverbs such as *hardly*, *merely*, and *simply*, modifies NP's with determiners that are **definite–** and **const+**, or that are **gen+**. This second class of adverbs can also modify NP's with *the* as a determiner. They do not modify NP's without determiners. The third class, exemplified by *almost*, *approximately*, and *relatively*, modifies the determiner itself. These adverbs are restricted to modifying **card+** determiners, as well as *all*, *double*, and *half*. The distinction between adverbs that modify NP's and ones that modify determiners can be seen in the NP's in (7) and (8).

(7) [Just][half the people]

(8) [Approximately half][the people]

This analysis of determiner adverbs is still somewhat preliminary, and has not yet been fully integrated with the feature system.

## 6 Future Work

There remains work to be done to fully capture all of the possible determiner constructions in English. Although constructions such as conjunctions are easily handled if overgeneration is allowed, blocking sequences such as *'one and some'* while allowing sequences such as *'one and all'* still remains to be worked out. Also, we believe that comparative constructions such as *more than* and *other than* will fall into our determiner phrase category, perhaps with trees similar in construction to the partitive and genitive NP trees.

There are still a handful of determiners that are not currently handled by our system. We do not have an analysis to handle *most*, *such*, *certain*, *other*, and *own*[5]. In addition, there are a set of lexical items that we consider adjectives (*enough*, *less*, *more*, and *much*) that have the property that they cannot occur with determiners[6]. We feel that a complete analysis of determiners should be able to account for this phenomena.

## 7 Conclusion

With this work, we present a syntactic account of determiner sequencing. Eight independently identified semantic features prove sufficient to account for the sequencing of a substantial portion of the English determiner system. The system handles single-word determiners and recursive constructions such as genitives and partitives, as well as determiner adverbs. Longer determiner sequences are constructed by composing any of the types. This work has been mostly implemented in an existing Lexicalized Tree Adjoining Grammar for English.

## 8 Acknowledgments

This work is part of an on-going project to develop a wide-coverage English grammar using FB-LTAG, and we would like to thank all of those involved with the project. We would like to give special thanks to people who have helped in specific ways on the determiner sequencing, particularly Christy Doran, B. Srinivas, Matthew Stone, and Dr. Aravind Joshi.

---

[5] The behavior of *own* is sufficiently unlike other determiners that it most likely needs a tree of its own, adjoining onto the right-hand side of genitive determiners.

[6] This is not the same set of lexical items that we counted as adjectives in the Introduction. Together, there are 14 lexical items marked as determiners in Collins that we feel should be adjectives.